\pdfoutput=1
\documentclass[10pt,twocolumn,letterpaper]{article}
\usepackage{wacv}
\usepackage{times}
\usepackage{epsfig}
\usepackage{graphicx}
\usepackage{amsmath}
\usepackage{amssymb}
\usepackage{bm}
\usepackage{xcolor}
\usepackage[ruled,vlined]{algorithm2e}
\usepackage{pifont}
\newcommand{\cmark}{\ding{51}}%
\newcommand{\xmark}{\ding{55}}%

\usepackage{float}
\usepackage{authblk}
\makeatletter
\renewcommand\AB@affilsepx{ , \protect\Affilfont}
\makeatother
\def\wacvPaperID{572} 

\wacvfinalcopy 

\ifwacvfinal
\def\assignedStartPage{9876} 
\fi
\ifwacvfinal
\usepackage[breaklinks=true,bookmarks=false]{hyperref}
\else
\usepackage[pagebackref=true,breaklinks=true,colorlinks,bookmarks=false]{hyperref}
\fi

\ifwacvfinal
\else
\fi

\begin{document}

\title{MVHM: A Large-Scale Multi-View Hand Mesh Benchmark for\\ Accurate 3D Hand Pose Estimation}


\author[1]{Liangjian Chen}
\author[2]{Shih-Yao Lin}
\newcommand\CoAuthorMark{\footnotemark[\arabic{footnote}]} 
\author[3]{Yusheng Xie\protect\CoAuthorMark\thanks{\textit{Work done outside of Amazon.}}}
\author[4]{Yen-Yu Lin}
\author[1]{Xiaohui Xie}
\affil[1]{University of California, Irvine}
\affil[2]{Tencent America}
\affil[3]{Amazon}
\affil[4]{National Chiao Tung University}
\affil[ ]{\textit {\{liangjc2,xhx\}@ics.uci.edu}}
\affil[ ]{\textit {mike.lin@ieee.org}}
\affil[ ]{\textit {yushx@amazon.com}}
\affil[ ]{\textit {lin@cs.nctu.edu.tw }}

\maketitle
\def\eg{\emph{e.g}\onedot} \def\Eg{\emph{E.g}\onedot}
\vspace*{-1.2cm}
\begin{abstract}
\vspace{-.3cm}
Estimating 3D hand poses from a single RGB image is challenging because depth ambiguity leads the problem ill-posed. 
Training hand pose estimators with 3D hand mesh annotations and multi-view images often results in significant performance gains. 
However, existing multi-view datasets are relatively small with hand joints annotated by off-the-shelf trackers or automated through model predictions, both of which may be inaccurate and can introduce biases. 
Collecting a large-scale multi-view 3D hand pose images with accurate mesh and joint annotations is valuable but strenuous.
In this paper, we design a spin match algorithm that enables a rigid mesh model matching with any target mesh ground truth.
Based on the match algorithm, we propose an efficient pipeline to generate a large-scale multi-view hand mesh (MVHM) dataset with accurate 3D hand mesh and joint labels.
We further present a multi-view hand pose estimation approach to verify that training a hand pose estimator with our generated dataset greatly enhances the performance.
Experimental results show that our approach achieves the performance of 0.990 in $\text{AUC}_{\text{20-50}}$ on the MHP dataset compared to the previous state-of-the-art of 0.939 on this dataset. 
%
Our datasset is available at~\href{https://github.com/Kuzphi/MVHM}{\color{blue}{https://github.com/Kuzphi/MVHM}}.

\end{abstract}

\begin{figure}
\centering 
\begin{tabular}{c}
\hspace{-0.5cm}
\includegraphics[width=0.46\textwidth]{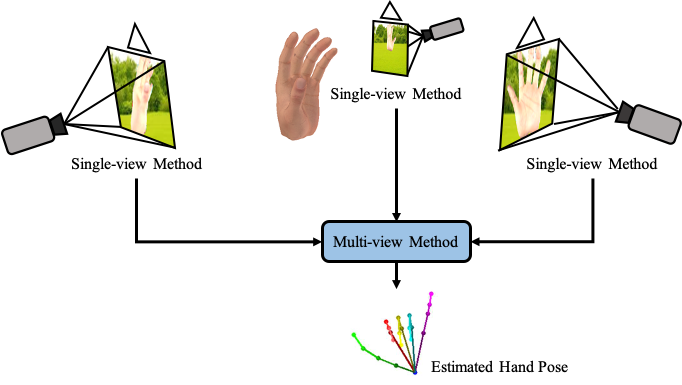}
\end{tabular}
\label{fig:1}
\caption{Our core idea. We build a synthetic dataset from a multi-view perspective, \eg, rendering hand images from different a angles. With the aid of this dataset, a single-view method takes the image from each view and generates a possible hand pose candidate. We proposed a multi-view method takes different single-view predictions as input and predicts the final result.}
\vspace{-.6cm}
\end{figure}
\vspace{-.4cm}

\section{Introduction}
\label{sec:intro}
%

Estimating 3D hand poses from images has attracted increasing attention because it is essential to a wide range of applications such as human-computer interaction (HCI)~\cite{anwar2019hand, lin2013airtouch,mikeicpr10}, virtual reality (VR)~\cite{cote2019deep, hung2016re}, augmented reality (AR)~\cite{chalasani2019simultaneous}, medical diagnosis~\cite{ismail2019hand}, and sign language understanding~\cite{zafrulla2011american}.
Although extensive research efforts have been made on this research topic for decades, there are still several unsolved challenges.
One of the most crucial challenges is to handle the issue of depth ambiguity present in single view 3D hand pose estimation.

Conventional studies mainly focus on inferring 3D hand poses from either depth or RGB images directly.
To address the problems caused by depth ambiguity,
Some previous studies~\cite{cai2018weakly,iqbal2018hand, chen2020dggan, chen_wacv21_temporal} want to address this problem by leveraging depthmap information. These works come up with several ways to introduce depthmap into the training procedure, such as making depthmap as intermediate supervision~\cite{iqbal2018hand} or using depth regularizer~\cite{cai2018weakly}.
%
%
On the other hand, recent studies~\cite{kokic2019learning,zimmermann2019freihand} point out that imposing 3D hand shape (mesh) supervision can boost both the performance of 3D hand pose and shape estimators.
It is clear that 3D hand shape brings richer hand structure information than hand keypoints.
%
%
Furthermore, a preset mesh serves as a strong prior to reduce the freedom of the hand, therefore mitigating depth ambiguity.
%
%
Along this line, several methods such as \cite{boukhayma20193d,baek2019pushing,zhang2019end,kulon2020weakly,yang2020seqhand, mm-hand,zhao2020image,zhao2020topk,zhao2020improved} are proposed.
Despite the potential, the aforementioned methods highly rely on a preset hand model learned with a large number of accurate 3D mesh annotations. 
Hence, a large-scale dataset with accurate annotations of mesh vertices is in great demand.
%

%
%
%
%

Accurate mesh ground truth is hard to be manually annotated in general.
The hand mesh annotations in most existing datasets are often annotated by hand shape estimator which can be inaccurate because hand mesh estimation itself is an even more challenging task.
Most existing methods leveraging mesh information for 3D hand pose estimation are derived based on a single view. 
However, mere mesh information is insufficient to address depth ambiguity.
Thus, 3D pose estimation still remains ill-posed in these methods.
%

%
%

The issue of depth ambiguity can be tackled by multi-view vision according to epipolar geometry.
Multi-view sensing systems can capture hand images from cameras in different angles and therefore depth information of hands can be accurately inferred as long as camera parameters are known.
Inspired by the above observations, we aim to build a large-scale multi-view hand mesh dataset that provides hand meshes and multi-view hand images simultaneously for training pose estimators.

%
%
%
%

In this work, we present an effective mechanism to synthesize 3D hand joint and mesh annotations, and establish a large-scale multi-view hand mesh (MVHM) dataset.
We acquired a hand mesh model with a rigging system, and 3D hand ground truth from existing datasets, and a rigged hand model to match the given ground truth to perform various gestures.
We render the hand model from different angles to collect multi-view images, as well as the 3D keypoints and mesh annotations to built MVHM.
Then, we determine if the generated MVHM dataset can be used to improve 3D hand pose estimators.
To this end, a multi-view based approach is developed for inferring 3D hand poses. 
The experimental results show that the resultant pose estimator can be greatly boosted by leveraging the generated MVHM dataset, and performs favorably against existing methods.
This work makes three major contributions, which are summarized as follows:
\begin{enumerate}
    \vspace{-.2cm}
    \item We propose an effective mechanism for compiling a large-scale multi-view hand mesh (MVHM) dataset for 3D hand pose estimator training. To the best of our knowledge, this is the first large-scale hand dataset with multi-view hand images, accurate mesh annotations, hand joint keypoints labels.
    \vspace{-.2cm}
    \item We present a multi-view hand pose estimation approach based on an end-to-end trainable graph convolutional neural network where information from multi-view images is utilized to predict 3D hand poses.
    \vspace{-.2cm}
    \item Our proposed approach achieves the state-of-the-art performance on the benchmark, the MHP dataset, in both single-view and multi-view settings.
    \vspace{-.3cm}
\end{enumerate}


\begin{table*}[h]
    \centering
    \caption{Comparison between our dataset with publicly available datasets. {\tt Auto} in field {\tt Annotation} represents that the annotation is made by some algorithms and therefore may not be accurate.{\tt Mano} means the emsh annotaion is fitted by Mano Model }
    \begin{tabular}{r||cccccrcc}
    \hline
        Dataset & RGB & Depth & Image Type & Resolution & Annotation & Dataset Size & Multi-View & Mesh \\
    \hline
        ICVL~\cite{tang2014latent}&   \xmark & \cmark & real & 320 $\times$ 320 & tracking & 18K &\xmark&\xmark\\ 
        NYU~\cite{tompson14tog}& \xmark& \cmark & real & 648 $\times$ 480 & tracking & 243K &\xmark&\xmark\\
        MSRA ~\cite{sun2015cascaded}& \xmark&\cmark & real & 1920 $\times$ 1080 & tracking & 76K&\xmark&\xmark \\
        BigHand2.2M~\cite{yuan2017bighand2} & \xmark& \cmark & real & 640 $\times$ 480 & marker & 2.2M&\xmark&\xmark\\
        STB ~\cite{zhang20163d}& \cmark&  \cmark& real& 640 $\times$ 480 & manual& 36K&\xmark&\xmark\\
        RHP ~\cite{zb2017hand}& \cmark& \cmark& synthetic&640 $\times$ 480 & synthetic & 44K & \xmark & \xmark\\
        Dexter+Object~\cite{sridhar2016real} & \cmark& \cmark& real& 640 $\times$ 480& manual&3K&\xmark&\xmark \\
        EgoDexter~\cite{mueller2017real}& \cmark & \cmark& real& 640 $\times$ 480& manual&3K&\xmark&\xmark \\
        MHP~\cite{gomez2017large} & \cmark & \xmark & real &480 $\times$ 480 &auto& 80K & \cmark & \xmark \\
        FreiHand~\cite{zimmermann2019freihand} &\cmark & \xmark & real &224 $\times$ 224 & auto & 134K & \xmark & Mano \\
        InterHand ~\cite{mooninterhand2} &  \cmark& \xmark & real & 512 $\times$ 334 & auto & 2.2M & \cmark & Mano\\
        Youtube Hand ~\cite{kulon2020weakly} &\cmark & \xmark & real & 256$\times$256  & auto & 47K & \xmark & Mano\\
    \hline
        \textbf{Ours} & \cmark & \cmark & synthetic& 256$\times$256 & synthetic &320K & \cmark & \cmark\\
    \hline
    \end{tabular}
    \label{tab:dataset}
\end{table*}

\section{Related Work}


%
\subsection{RGB based 3D Hand Pose Estimation}
%
%
%
%

RGB cameras are much more widely used than depth sensors. 
Estimating 3D hand poses merely from monocular RGB images are more practical and active in the literature~\cite{cai2018weakly,chentaChen2018GeneratingRTgan,iqbal2018hand,mueller2018ganerated,tekin2019h+,yang2018disentangling,zb2017hand,kong2020sia,kong2019adaptive,chen2018generating}.
The pioneering work by Zimmermann and Brox~\cite{zb2017hand} utilizes convolutional neural networks (CNN) to extract image feature, and feed camera parameters with these features to a 3D lift network where depth information is then estimated.
Based on \cite{zb2017hand}, Iqbal~\etal \cite{iqbal2018hand} leverage depth maps as intermediate supervision.
Meanwhile, Cai~\etal~\cite{cai2018weakly} propose a weakly supervised approach that reconstructs the depth map and uses it as a regularizer during model training.

\subsection{3D Hand Mesh Estimation}
\label{rel:mesh}

%
%
%
%
%
%
%
%
%
%

3D hand pose estimation provides sparse joint locations. 
However, many computer vision applications would benefit more from hand shape information than sparse joints.
Therefore, 3D hand mesh estimation, an effective shape representation, has emerged as an increasingly popular topic~\cite{ge2019handshapepose,boukhayma20193d,baek2019pushing,joo2018total,zhang2019end}. 
Most methods~\cite{boukhayma20193d,baek2019pushing,zhang2019end,kulon2020weakly,yang2020seqhand} are developed around a pre-defined deformable hand mesh model called MANO~\cite{romero2017embodied}. 
Because of the high degree of freedom and complexity of the hand gesture, searching for the right hand mesh in such a high dimensional space is quite challenging.
Using this MANO model often relies on strong prior to constrain the model to only regress low-dimensional model parameters, and may ignore the high-dimensional information.
Ge~\etal~\cite{ge2019handshapepose} argue that mesh is a graph-structure data, and propose to directly regress 3D mesh vertices through graph convolutional neural network (GCN) with a pre-defined mesh graph.
%
%

\subsection{Multi-View Hand Pose Estimation}

Unlike single-view pose estimation, few research efforts focus on 3D hand pose estimation from multi-view data.
%
%
%
Ge~\etal~\cite{ge2016robust} first introduce multi-view CNN to formulate it as an estimation problem.
Their method assumes that hand joint locations independently follow 3D Gaussian distributions, and uses CNN to estimate the mean and variance of the location distribution of each joint.
The main drawbacks of their method include 1) its inability to train in an end-to-end manner and 2) its impractical assumption about the independence among different joint locations.
Simon~\etal~\cite{simon2017hand} propose a multi-view system which is trained to progressively improve hand keypoints detection. 
Their method would work well on fine-tuning a well pre-trained estimator but could not train a 3D hand pose estimator from scratch.

\subsection{3D Hand Pose Benchmark}

%

There exist extensive research efforts such as \cite{tang2014latent,tompson14tog,sun2015cascaded,yuan2017bighand2,zhang20163d,zb2017hand,sridhar2016real,mueller2017real,mooninterhand2,kulon2020weakly,zimmermann2019freihand} on building hand datasets for 3D hand pose estimation.
We summarize the publicly available hand datasets and our dataset in Table~\ref{tab:dataset}.
%
%
Most existing datasets do not contain mesh information, since labeling hand meshes manually is almost infeasible for human annotators.

%
%
%

To address the issue of labor-intensive annotations, recent studies \cite{zimmermann2019freihand,kulon2020weakly, chen_wacv21_temporal} propose semi-automatic ways to label RGB images. 
FreiHand (Zimmermann~\etal~\cite{zimmermann2019freihand}) use an iterative process where the trained models first make predictions on the images and then the annotators are asked to make necessary adjustments.
YoutubeHand (Kulon~\etal~\cite{kulon2020weakly}) run OpenPose~\cite{cao2018openpose} to get 2D annotations, upon which the parameters of the MANO model are regressed.
Thresholding according to confidence scores is applied to remove those with low confidence, and hence ensures annotation quality.
Despite the progress on efficiency and efficacy of labeling RGB images, the accuracy of annotation relies heavily on the pre-trained models used in the process. 
In addition, these methods rely on the MANO model as the ground-truth mesh generator, which could lose high-dimensional information of hands, as mentioned in Section~\ref{rel:mesh}. 
Compared to existing datasets, our dataset consists of large-scale RGB images and includes a variety of sequences. 
In addition, synthetic nature provides 100\% accurate annotation for both hand joints and mesh.
We make the first attempt to collect the dataset that provides large-scale, multi-view training images, thereby enhancing pose estimator training with a multi-view perspective.
\section{Generating Muti-View Dataset}

\begin{algorithm}
\SetAlgoLined
\KwIn{\\
{
$C$ is the array of 3D keypoints ground truth that we want our mesh to match with.\\
$B$ is the array of hand bones in the rig system. Each bone has two attributes, head and tail, which represent the beginning and end location of the bone.\\
$B$ and $C$ is stored in an array whose orders are shown in Figure \ref{fig:jointexample}(a) and \ref{fig:jointexample}(b)}
}
\Begin{
Move B[0].tail to location C[0] \;
\tcp{spin the bone inside the palm}
\For{$i \in \{1,5,9,13,17\}$ }{
    \tcp{bone vector}
    $u \leftarrow $ B[i].tail - B[i].head\;
    $v \leftarrow $ C[i] - C[0] \;
    adj $\leftarrow i + 4$\;
    \If {$i$ = $17$}{
        adj $\leftarrow i - 4$\;
    }
    \tcp{reference vector}
    $\text{ref}_{\text{ori}} \leftarrow $B[adj].tail - B[adj].head\;
    $\text{ref}_{\text{aft}} \leftarrow $C[adj] - C[0] \;
    Move B[i] to match groundturth\;
    \tcp{sign vector}
    ${e_1} = u \times \text{ref}_{\text{ori}}  \times u $\;
    ${e_2} = v \times \text{ref}_{\text{aft}}  \times v$ \;
    Spin B[i] with the angle between $e_1$ $\&$ $e_2$\;
}

\For{$i \leftarrow 1$ \KwTo $21$ }{
    \If {$i$ mod $4$ $\neq$  $1$}{
        Move B[i] to match groundturth\;
        B[i] performs the same spin as B[i-1] \;
    }
}
}
\caption{Spin matching Algorithm for rigging hand mesh base on 3D hand pose ground truth}
\label{algo:match}
\end{algorithm}
\begin{figure}
\centering 
\begin{tabular}{ccc}
\includegraphics[width=0.13\textwidth]{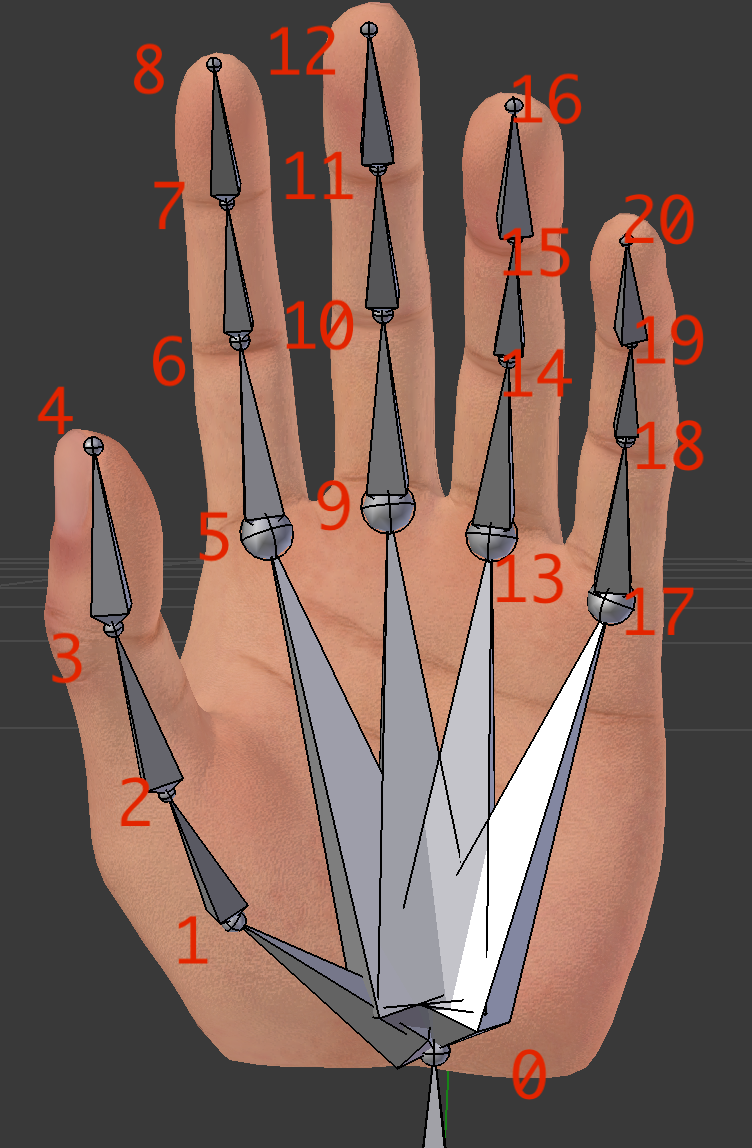}&
\includegraphics[width=0.13\textwidth]{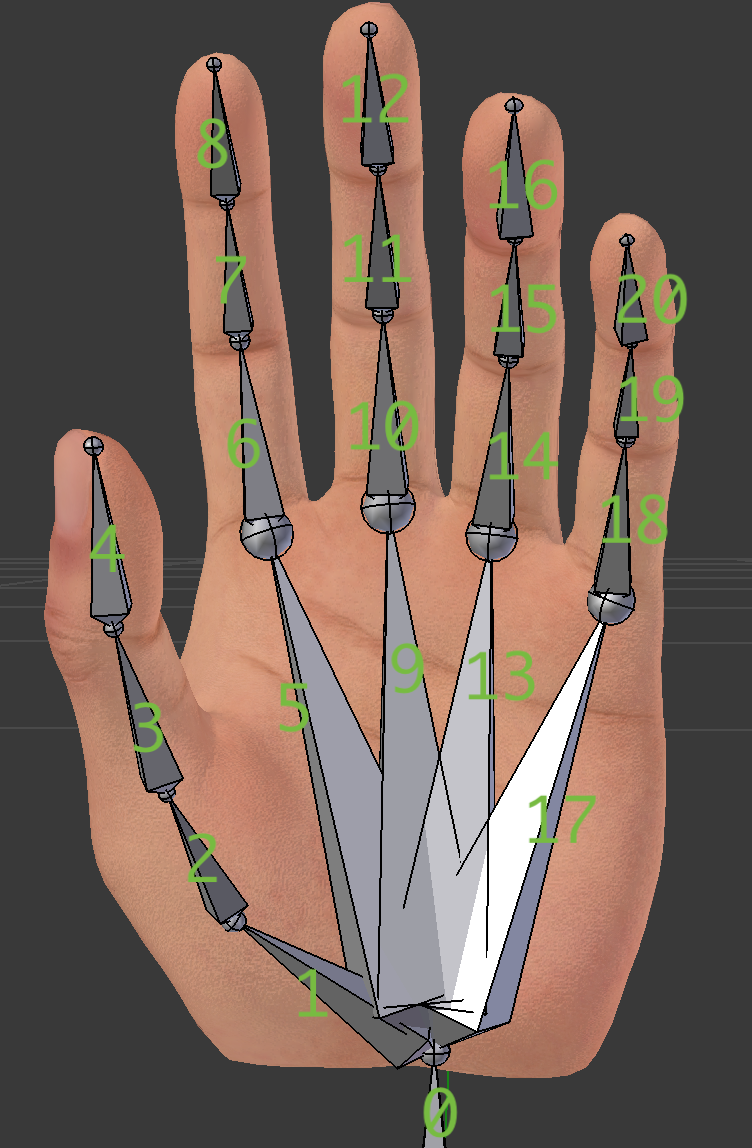}&
\includegraphics[width=0.13\textwidth]{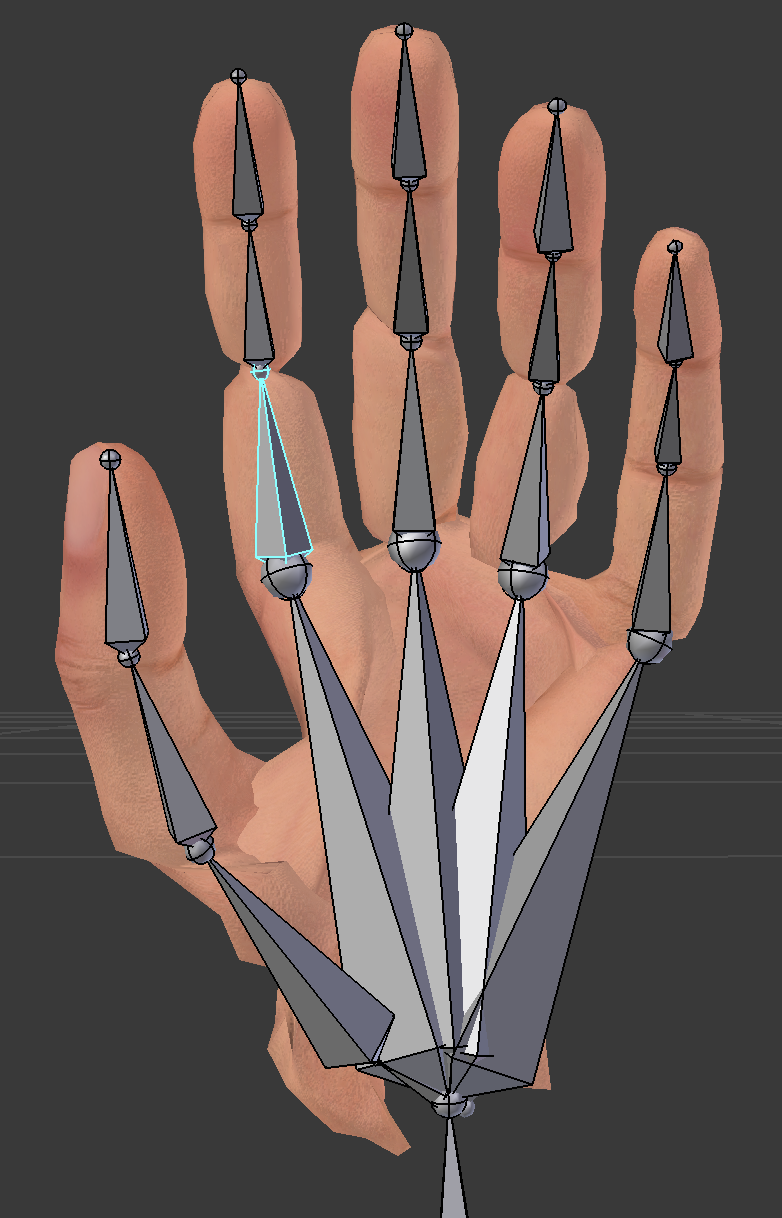}\\
(a)&(b)&(c)\\
\end{tabular}
\caption{
An example of hand joint and bone labels and their orders adopted in this work. (a) Joint labels. (b) Bone labels, which are used by Algorithm \ref{algo:match} during spin matching.
(c) A failure case when directly rigging the hand mesh based on the bone coordinates without using Algorithm \ref{algo:match}.
}\label{fig:jointexample}
\vspace{-.4cm}
\end{figure}

\begin{figure*}[t]
\begin{center}
	\begin{tabular}{cccccccccc}
	\includegraphics[width=0.095\textwidth]{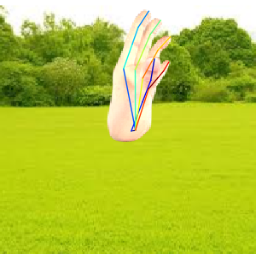}
	\includegraphics[width=0.095\textwidth]{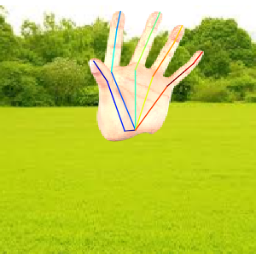}
	\includegraphics[width=0.095\textwidth]{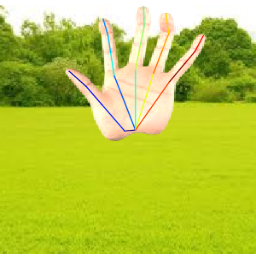}
	\includegraphics[width=0.095\textwidth]{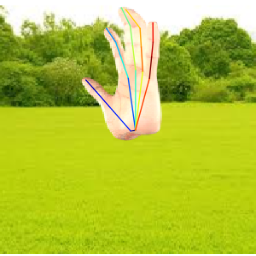}
	\includegraphics[width=0.095\textwidth]{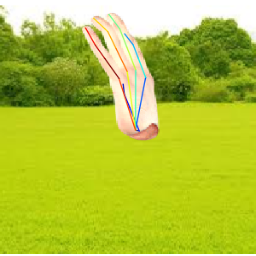}
	\includegraphics[width=0.095\textwidth]{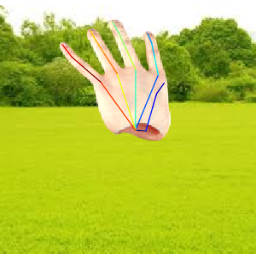}
	\includegraphics[width=0.095\textwidth]{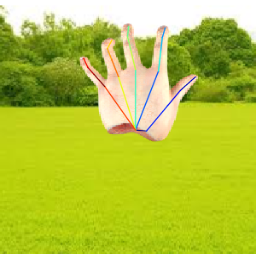}
	\includegraphics[width=0.095\textwidth]{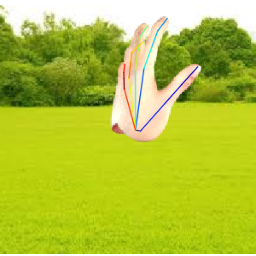}
	\includegraphics[width=0.095\textwidth]{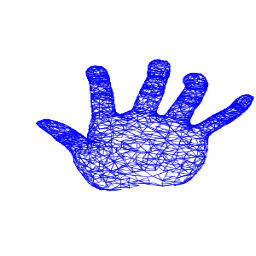}
	\includegraphics[width=0.095\textwidth]{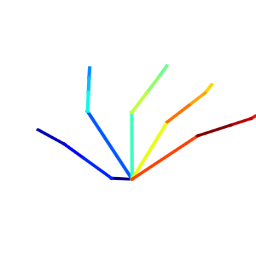}
	\end{tabular}
	\begin{tabular}{cccccccccc}
	\includegraphics[width=0.095\textwidth]{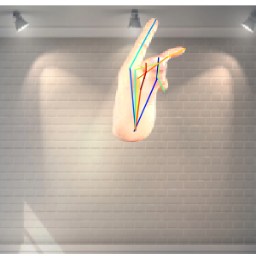}
	\includegraphics[width=0.095\textwidth]{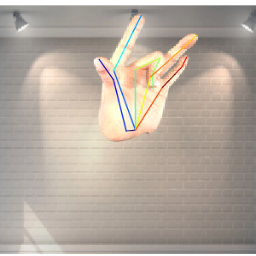}
	\includegraphics[width=0.095\textwidth]{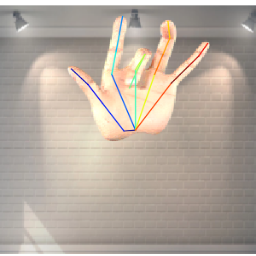}
	\includegraphics[width=0.095\textwidth]{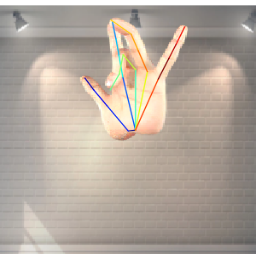}
	\includegraphics[width=0.095\textwidth]{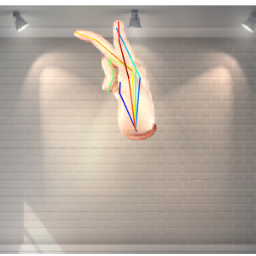}
	\includegraphics[width=0.095\textwidth]{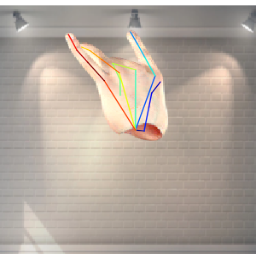}
	\includegraphics[width=0.095\textwidth]{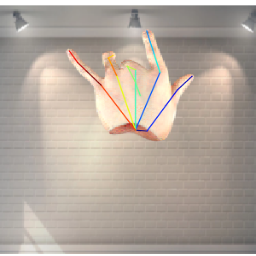}
	\includegraphics[width=0.095\textwidth]{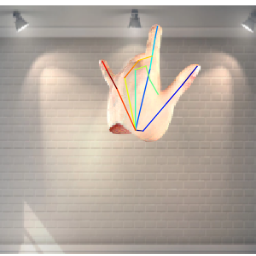}
	\includegraphics[width=0.095\textwidth]{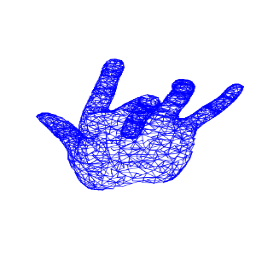}
	\includegraphics[width=0.095\textwidth]{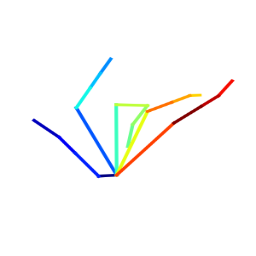}
	\end{tabular}
	\begin{tabular}{cccccccccc}
	\includegraphics[width=0.095\textwidth]{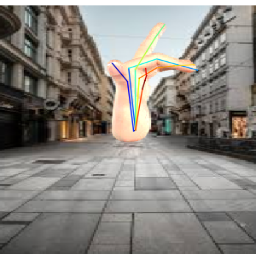}
	\includegraphics[width=0.095\textwidth]{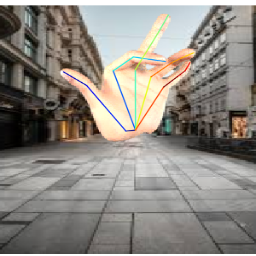}
	\includegraphics[width=0.095\textwidth]{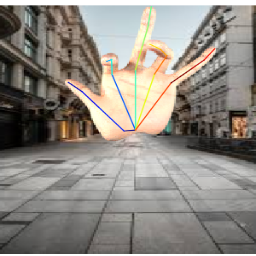}
	\includegraphics[width=0.095\textwidth]{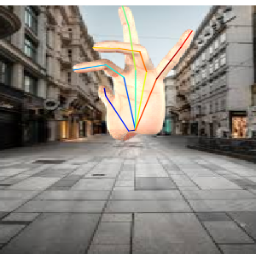}
	\includegraphics[width=0.095\textwidth]{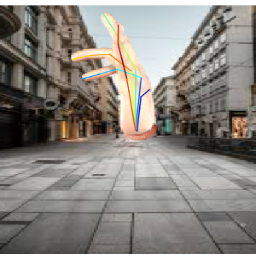}
	\includegraphics[width=0.095\textwidth]{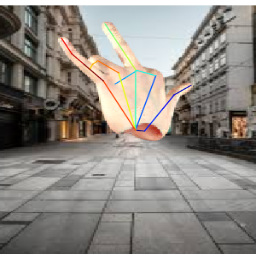}
	\includegraphics[width=0.095\textwidth]{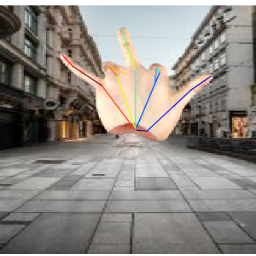}
	\includegraphics[width=0.095\textwidth]{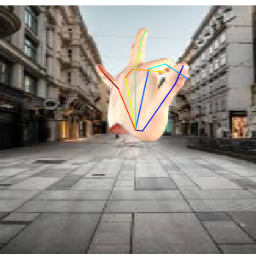}
	\includegraphics[width=0.095\textwidth]{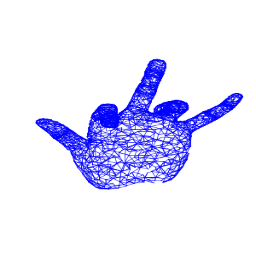}
	\includegraphics[width=0.095\textwidth]{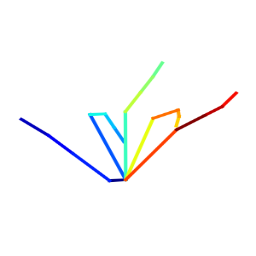}
	\end{tabular}
\end{center}
\caption{Some examples of the MVHM dataset. Column 1 to Column 8 shows the image with 2D annotation from view 1 to view 8, Column 9 shows the mesh. Column 10 shows the 3D annotation}
\label{fig:MVHM}
\vspace{-.3cm}
\end{figure*}

%
%
%

%
Currently, there exists no dataset providing \emph{both} large-scale mesh and multi-view annotations of 3D hands, although many potential applications can benefit from such a dataset.
Therefore, we create a new dataset called Multi-View Hand Mesh (MVHM) dataset for training multi-view hand pose estimators.
To get accurate mesh and joint annotation, we use a well-made hand model from TurboSquid\footnote{\url{https://www.turbosquid.com}} which provides around 2000 mesh vertices as well as an armature system to form various hand gestures. We render the images in the open-source software Blender\footnote{\url{https://www.blender.org}}. 
%

%
%
%

%
In order to generate images and meshes from different gestures, we deploy the NYU dataset~\cite{tompson14tog} which provides various hand pose and accurate keypoints annotation and rigged hand bone in our model to match with the given groundtruth.
The hand bone in the rigged system consists of $7$ degrees of freedom, $3$ for its bone head, $3$ for its bone tail, and $1$ for its spin. The first $6$ degrees of freedom determine the location of the bone and the last one represents its orientation. 
In order to rig the hand joints and the mesh surface correctly, we need to consider both location and orientation.
Figure~\ref{fig:jointexample}(c) shows an example of a distorted mesh obtained when we do not perform orientation match but simply move the bone location.

For this purpose, we define a bone vector as the difference between the bone tail and head. Assume $u$ as the bone vector of the current bone we are working on. We take its adjacent bone's vector as the spin reference $ref$.
We define spin sign vector as $u \times ref \times u$, and make sure this vector does not change after matching bone with groundturth.
The detailed algorithm is summarized in Algorithm~\ref{algo:match}.
%


%
%
%

\begin{figure}
\centering 
\begin{tabular}{c}
\includegraphics[width=0.42\textwidth]{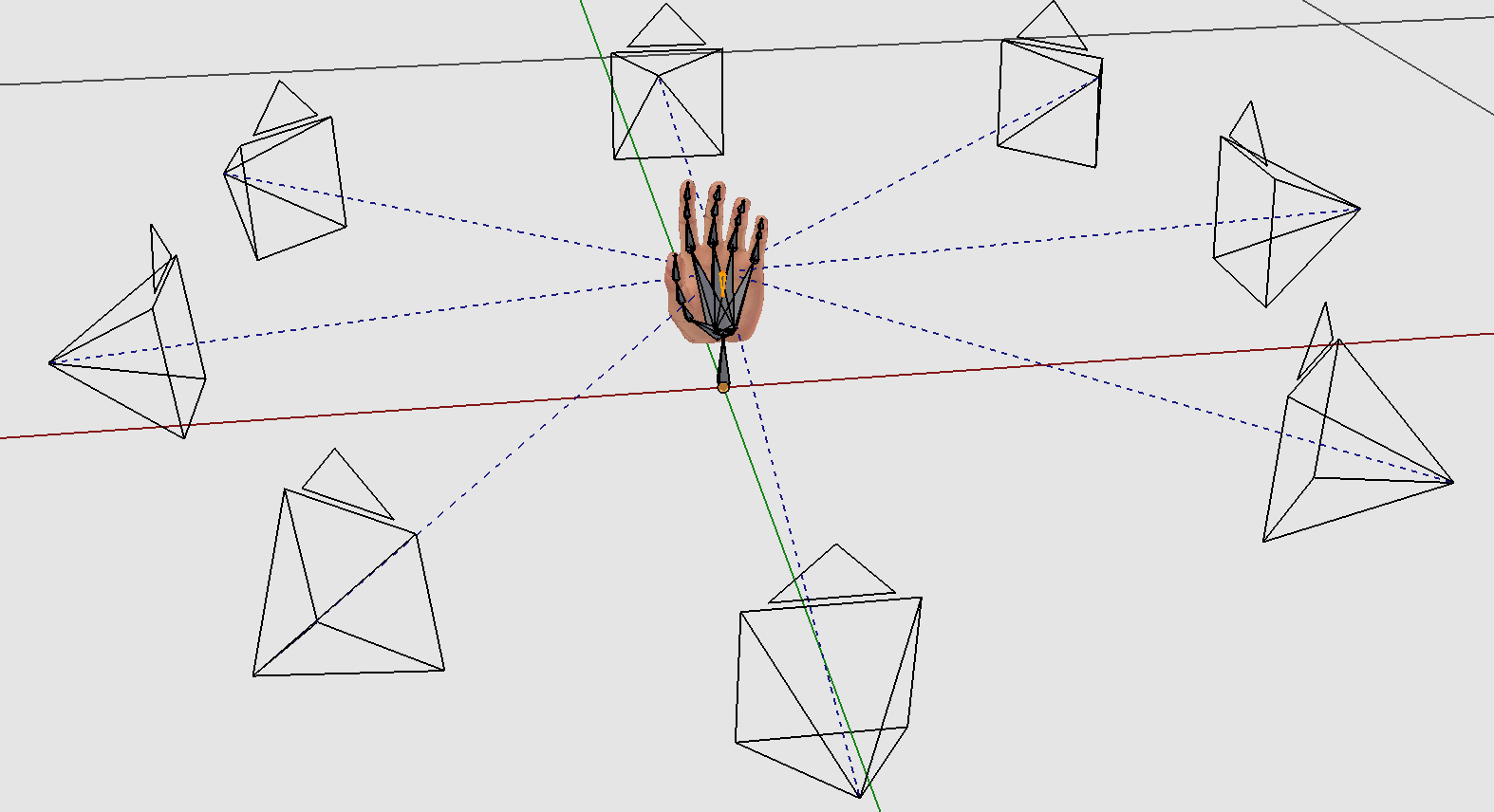}
\end{tabular}
\caption{
Synthetic scene when rendering hands in MVHM.
}\label{fig:scene}
\vspace{-.3cm}
\end{figure}

For each ground-truth gesture, we set 8 different camera positions that are evenly located on a circle within the plane perpendicular to the palm.
All 8 cameras point to the center of the palm to ensure that the hand locates at the center of each rendered image. 
Figure~\ref{fig:scene} shows the scene when we render hand images.
%

%
%

To increase the diversity of the collected MVHM, we randomly change the intensity of the light and global illumination in the blender.
In addition, we select some background scenes from online sources, and randomly use them as background during our rendering.
We render $320,000$ images of resolution $256\times 256$ for MVHM construction. 

%
%
%
%

We emphasize that each sample in MVHM comes with full annotations of 21 hand joint and 2651 mesh vertices.
Following the setting in \cite{zb2017hand}, each finger is fully represented by 4 keypoints (Metacarpophalangeal, Proximal interphalangeal, Distal interphalangeal and fingertip), additionally. 
Carpometacarpal joints are also labeled in MVHM. 
Figure~\ref{fig:jointexample}(a) shows a sample of the hand joint labels.
In addition, we also release the hand segmentation mask, the camera intrinsic matrix, and the optical flow for each sample.
Nevertheless, we in this paper only use the multi-view and mesh information from the collected MVHM dataset for hand pose estimator training.

\section{Methodology}


\subsection{Overview}

%
%
%

Given an RGB image of a hand $\mathbf{I} \in \mathbb{R}^{W \times H \times 3}$, our goal is to estimate the $3$D joint locations of the hand $\mathbf{P}_{j} \in \mathbb{R} ^ {k \times 3}$, where $W$ and $H$ denote the image height and width respectively, and $K$ is the number of the hand joints.
Recent studies~\cite{ge2019handshapepose,boukhayma20193d} have demonstrated that using the mesh distance loss as an intermediate supervision during training can boost the performance of the learned hand pose estimator.
Inspired by the approach~\cite{ge2019handshapepose}, we define a hand mesh as a bidirectional graph $\mathbf{G(V, \Lambda)}$, where $\mathbf{V}$ is the vertex set and $\mathbf{\Lambda}$ is the adjacency matrix.
We also assume that $\mathbf{V}$ contains $N$ different elements (\ie, points on the mesh) and our mesh estimator would predict the 3D locations $\mathbf{P}_{m} \in \mathbb{R} ^ {N \times 3}$ for all vertices in $\mathbf{V}$.

%
%
%

\begin{figure}[t]
\centering 
\includegraphics[width=0.3\textwidth]{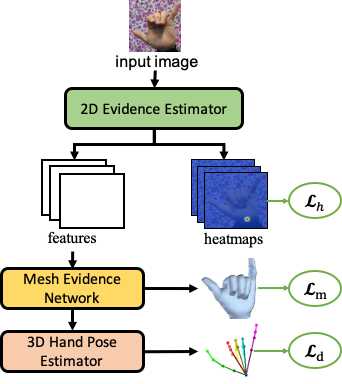}
\caption{
Overview of our single-view method. Given a single-view RGB image, the 2D evidence network predicts its heatmap and outputs the encoded image features. 
The mesh evidence network takes image features as input and outputs the hand mesh. 
Based on the estimated mesh, the 3D pose estimator gives final hand pose prediction.
%
}
\label{fig:single}
\vspace{-0.3cm}
\end{figure}

In our single-view approach, we use the stacked hourglass~\cite{newell2016stacked} as the CNN backbone to extract hand features from an image.
The graph convolution network (GCN) is applied to estimate the 3D pose and mesh.
Figure~\ref{fig:single} shows the architecture of our single-view network, which consists of three major components: the 2D evidence network, mesh evidence network, and 3D pose estimator. 
These components are elaborated in the following subsections.

\begin{figure*}[h]
\centering 
\includegraphics[width=0.85\textwidth]{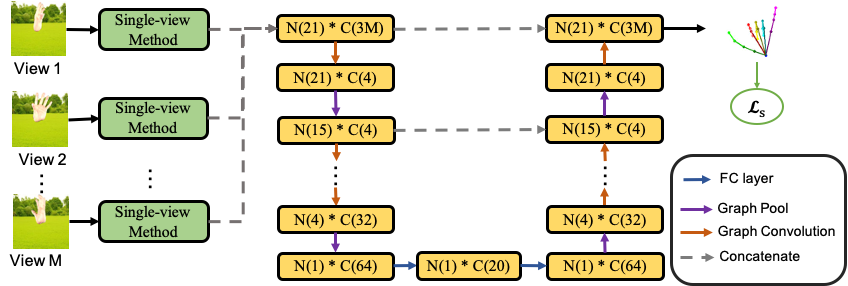}
\caption{
Overview of the multi-view method. 
%
The single-view method first predicts the hand pose for each view independently. 
A graph U-Net takes the concatenation of these single-view predictions as input, and estimates the final pose estimation. 
$N(\cdot)$ and $C(\cdot)$ represent the number of nodes in the graph and feature size of each node, respectively. 
%
}
\label{fig:multi}
\vspace{-0.2cm}
\end{figure*}

\subsection{2D Evidence Network}

%
%
%
%
%
%

The 2D evidence network offers two main functionalities. 
First, it estimates hand keypoint heatmaps to obtain the 2D hand joint locations.
Second, it extracts image features that then serve as the input to the mesh evidence network. 
We denote the estimated heatmaps as $\bm{H} \in \mathbb{R}^{K\times H\times W}$.
As shown in Figure~\ref{fig:single}, the hourglass backbone gives two outputs, including the estimated hand joint heatmaps and the extracted features.
The ground-truth heatmaps $\bm{\bar{H}}_s$ are obtained by smoothing the keypoint location $k^{th}$ with Gaussian blur.
To train the 2D evidence network, we apply the heatmap loss $\bm{\mathcal{L}}_{h}$ to each hourglass block as supervision.
The heatmap loss is defined by
\begin{equation}
\label{eq:heatmaploss}
    \bm{\mathcal{L}}_{h} =\frac{1}{S * K} \sum_{s = 1}^S \sum_{k = 1}^K||\bm{H}_k^s - \bm{\bar{H}}_k||_F^2,
\end{equation}
where $S$ and $K$ denote the number of the hourglass blocks and keypoints, respectively.

\subsubsection{Mesh Evidence Network}

%
%
%

%

Our mesh evidence network is built on the basis of spectral GCN~\cite{bruna2013spectral}. 
Given the image features extracted by the 2D evidence network, our mesh evidence network estimates the 3D hand mesh.
A 3D hand mesh is represented by a set of vertex coordinates $\mathbf{P}_{m} \in \mathbb{R} ^ {N \times 3}$ where $N$ is the number of the vertices in the hand mesh.
We represent a hand mesh as a graph $\mathbf{G(V, \Lambda)}$, where $\mathbf{V}$ is the vertices set, and $\mathbf{\Lambda}$ is the adjacency matrix.
$\mathbf{\Lambda}_{i,j}$ is $1$ if there is an edge between vertex $i$ and vertex $j$, otherwise it is $0$.

%

Specifically, we first normalize the adjacency matrix $\mathbf{\Lambda}$ via graph Laplacian operation and obtain a normalized adjacency matrix $\bm{L} = \bm{I} - \bm{D}^{-\frac{1}{2}}\Lambda\bm{D}^{-\frac{1}{2}}$, where $\bm{D}$ is the degree matrix of graph $\mathbf{G}$  and $\bm{I}$ is an identity matrix.  
Graph spectral decomposition is then used to decompose the normalized adjacency matrix $\bm{L}$ as $\bm{UAU}^T$, where $\bm{A} = diag(\lambda_1, \lambda_2,... , \lambda_N)$ consists of the eigenvalues of the $\bm{L}$, where $\lambda_{max}$ is the largest eigenvalue of $\bm{L}$. 


Following~\cite{defferrard2016convolutional}, we define the convolution kernel $\bm{\hat{A}}$ in GCN as
\begin{equation}
\bm{\hat{A}} = \\diag(\sum_{i=0}^S\alpha_i\lambda_1^i, ... ,\sum_{i=0}^S\alpha_i\lambda_C^i),
\end{equation}
where $\alpha$ is the kernel parameter and $S$ is a pre-set hyper-parameter used to control the receptive field.

%

Thus, the GCN convolutional operation is defined by
\begin{equation}
\bm{F'} = \bm{U\hat{\Lambda} U}^T\bm{F\theta}_i = \sum_{i=0}^{S} \alpha_i\bm{L}^i\bm{F\theta}_i,
\end{equation}
where $\bm{F}\in \mathbb{R}^{N\times F_{\text{in}}}$ and $\bm{F}' \in \mathbb{R}^{N\times F_\text{out}}$ indicate the input and output features respectively,  and $\bm{\theta}_i \in \mathbb{R}^{F_\text{in} \times F_\text{out}}$ is trainable parameter used to refine the input feature and control the output channel size.

%
%

Since our hand mesh surface is composed of $2561$ vertices, it takes a huge computational cost to apply the above operation to each vertex because the time complexity of matrix multiplication for $\bm{L}^i$ is $O(N^3)$.
Therefore, we utilize the Chebyshev polynomial approximation to reduce the complexity. 
The convolutional operation is then defined by
\begin{equation}
\bm{F'} = \\ \sum_{i=0}^{S} \alpha_iT_i(\hat{\bm{L}})\bm{\theta}_i,
\end{equation}
where $T_k(x)$ is the $k^{th}$ Chebyshev polynomial and $\hat{\bm{L}} = 2\bm{L} / \lambda_{max}- \bm{I}$ is used to normalize the input features.


%
%
%

To enable our model to learn both local and global features, we adopt a scheme that is used in~\cite{defferrard2016convolutional,ge2019handshapepose} for generating hand meshes from coarse to fine. 
We leverage the heavy-edge matching algorithm to coarsen the graph by three different coarsening levels, and record the mapping between graph nodes in every two consecutive levels.
In the forward pass, our model first constructs the most coarse hand mesh and then up-samples more nodes from the coarse graph to the fine graph based on the stored mappings.
%


At the last layer of the GCN, we set $F_\text{out}$ to $3$ to represent the 3D coordinate vertices. 
Also, we apply the $l_2$ loss between the ground-truth mesh and prediction map as the mesh loss function:
\begin{equation}
\label{eq:meshloss}
    \bm{\mathcal{L}}_{m} =\frac{1}{N} ||\bm{P_{m}} - \bm{\bar{P}_{m}}||_F^2.
\end{equation}

\subsection{3D Depth Evidence Network}

%
%

%

The proposed 3D evidence network infers the depth of 3D hand keypoints $P_d$ from the predicted hand mesh $P_m$ by the mesh evidence network.
Taking $P_m$ as the input, we adopt a two layers GCN with a similar structure of the mesh evidence network to predict the pose features. 
These pose features are then fed to two fully connected layers to regress the depth of 3D hand keypoint locations.
The corresponding loss is defined by  
\begin{equation}
\label{eq:depthloss}
    \bm{\mathcal{L}}_{d} =\frac{1}{K} ||\bm{D} - \bm{\bar{D}}||_F^2,
\end{equation}
where $\bm{D}\in \mathbb{R}^{K}$ and $\bm{\bar{D}} \in \mathbb{R}^{K}$ represent the predicted and the ground-truth joint depths, respectively.
%

%
%

To infer the 3D hand keypoints, we use non-maximum suppression to get the 2D coordinates from the estimated heatmaps. 
With the estimated 2D coordinates and the depth map calculated by the 3D depth evidence network, we then obtain 3D coordinates in the camera coordinate system. 
Since the camera parameters are known, we are then able to infer hand keypoints in the world system.

\subsection{Multi-view Method}

%
%

Based on our single-view method, we propose a simple yet effective multi-view approach to hand pose estimation. 
Figure \ref{fig:multi} illustrates the core idea of our approach.
%

%

Our single-view method predicts the 3D hand pose for each view independently.
We concatenate these view-specific predictions on their coordinate channel. 
The concatenated prediction serves as the input features to a graph U-Nets\cite{gao2019graph} and predicts the final 3D hand keypoints.
We utilize the $L_2$ distance as the loss function in our multi-view network, \ie,
\begin{equation}
\label{eq:poseloss}
    \bm{\mathcal{L}}_{s} =\frac{1}{K} ||\bm{P}_j - \bm{\bar{P}}_j||_F^2,
\end{equation}
where $\bm{P}_j$ and $\bm{\bar{P}}_j$ represent the predicted and the ground-truth joint depth, respectively.
\section{Experiment Setting}

\begin{figure}[t]
\begin{center}
	\begin{tabular}{ccccc}
		\includegraphics[width=0.085\textwidth]{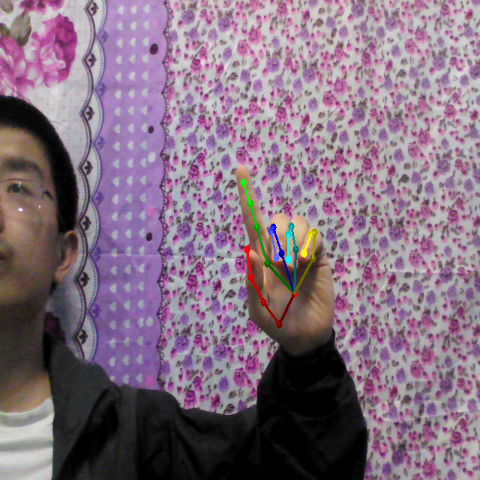}
		\includegraphics[width=0.085\textwidth]{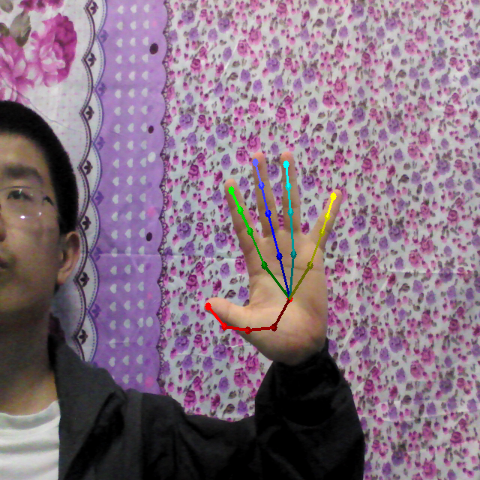}
		\includegraphics[width=0.085\textwidth]{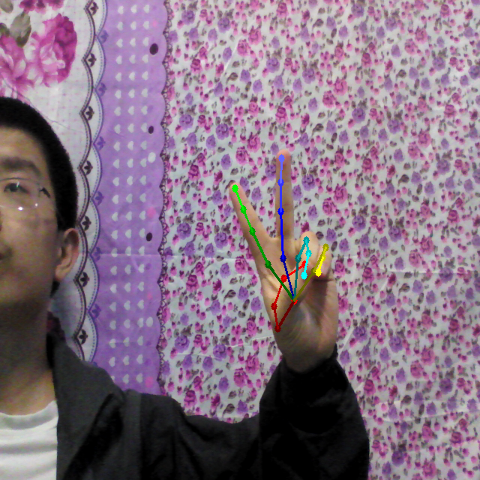}
		\includegraphics[width=0.085\textwidth]{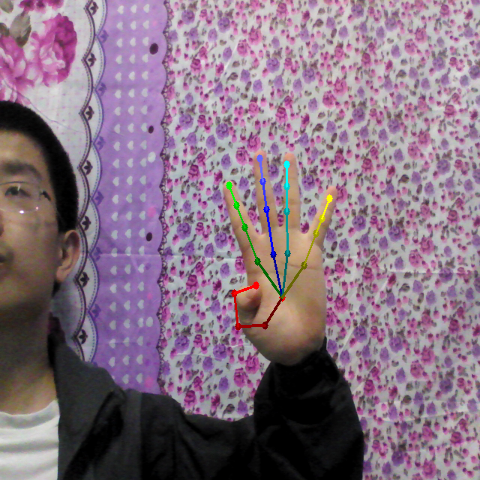}
		\includegraphics[width=0.085\textwidth]{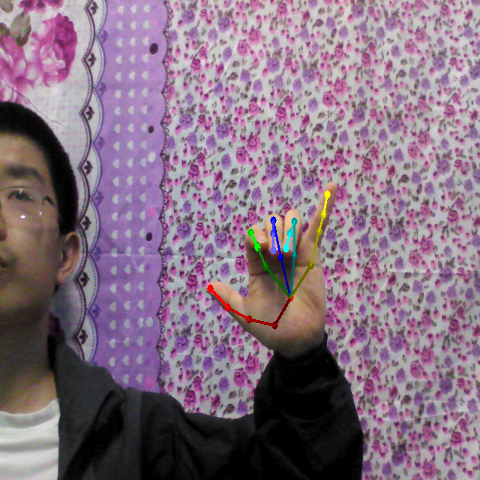}
	\end{tabular}
	\begin{tabular}{ccccc}
		\includegraphics[width=0.085\textwidth]{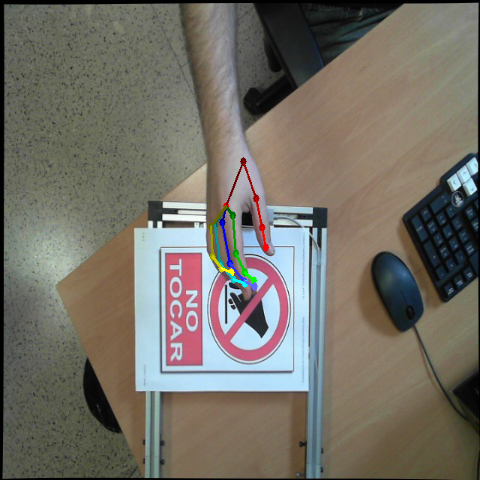}
		\includegraphics[width=0.085\textwidth]{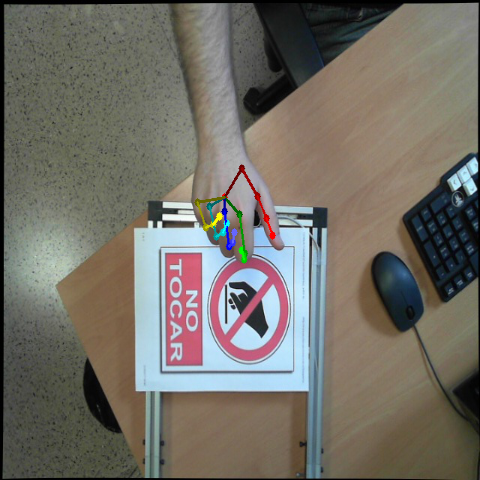}
		\includegraphics[width=0.085\textwidth]{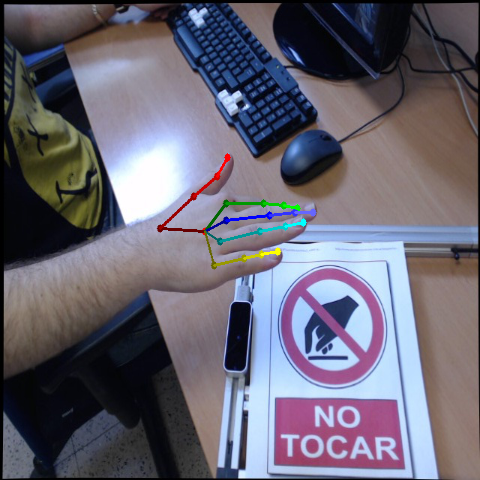}
		\includegraphics[width=0.085\textwidth]{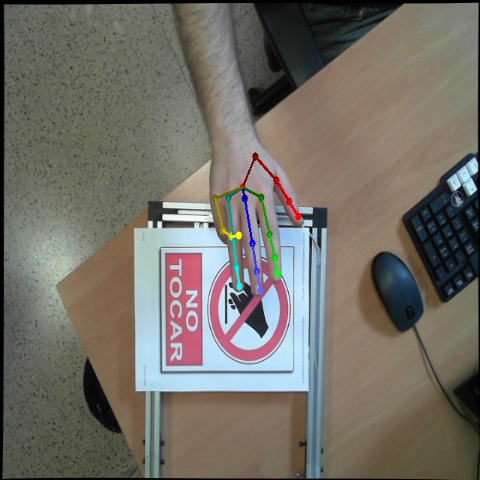}
		\includegraphics[width=0.085\textwidth]{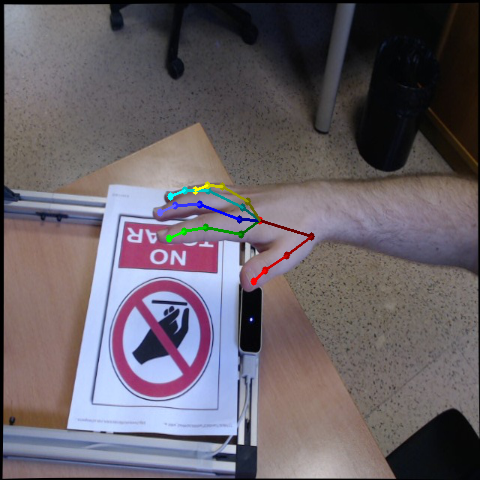}
	\end{tabular}
\end{center}
\caption{
Examples of the two hand pose datasets used for evaluation.
The first row shows images with the annotated hand poses from the STB dataset~\cite{zhang20163d} while the second row shows those from the MHP dataset~\cite{gomez2017large}. 
}
\label{fig:dataset}
\vspace{-0.1cm}
\end{figure}



\subsection{Datasets for Evaluation}

We evaluate our single-view approach on two benchmark hand pose datasets, including the Stereo Tracking Benchmark (STB) Dataset~\cite{zhang20163d} and the Multi-view 3D Hand Pose (MHP) dataset~\cite{gomez2017large}.
The proposed multi-view approach is evaluated on the MHP dataset.
Both the MHP and STB datasets provide real hand video sequences performed by different people in various backgrounds. 
The hand joint annotations of the STB dataset are manually labeled while the annotations of the MHP dataset are obtained by using the Leap Motion sensor.
The MVHM dataset we build is used in all of our experiments.
We aim at determining if training the hand pose estimators with the MVHM dataset can be effectively improved in different experimental settings.

For the STB dataset, we use its SK subset, which contains 6 different hand videos, to evaluate our approach.
Following the train-validation split setting in~\cite{ge2019handshapepose}, we take the first video as the validation set while the rest videos serve as the training set.

The MHP dataset includes 21 different hand motion videos.
Each hand motion video provides hand RGB images and multiple types of annotations in each sample, including bounding boxes and 2D/3D hand joint locations.
Figure~\ref{fig:dataset} displays some examples of the STB and MHP datasets.
We follow~\cite{cai2018weakly,zb2017hand} and apply the standard data pre-processing for both of the STB and MHP datasets.
During data pre-processing, we firstly crop the images to remove the irrelevant background and make sure the hands are located at the center of the images.
All the cropped images are then resized to resolution $256\times256$. 
%
Secondly, we follow the mechanism used in~\cite{cai2018weakly} to change the hand center from the palm center to the joint of the wrist for data in both of the STB and MHP datasets.

\begin{table}
\centering
\caption{
Ablation studies of $3$D hand pose estimation on the STB and MHP datasets. $\uparrow$: higher is better; $\downarrow$: lower is better; The measuring unit of EPE is millimeter(mm). SV stands for the single-view method and MV represents the multi-view method.
}
    \begin{tabular}{lcccc}
    \hline
&$\text{AUC}_\text{0-50}\uparrow$ &$\text{AUC}_{\text{20-50}}\uparrow$ & $\text{EPE}_m\downarrow$ \\
\hline
MHP Dataset&&& \\
\hline
SV w/o MVHM   &0.604 &0.802  &22.13 \\
SV w/ MVHM    &0.660 &0.857  &18.09 \\
MV w/o MVHM   &0.832 &0.985  & 8.43 \\
\textbf{MV w/ MVHM}     &\textbf{0.895} &\textbf{0.990}  & \textbf{5.20} \\
\hline
\hline
STB Dataset&&& \\
\hline
SV w/o MVHM   &0.820 &0.987  & 8.95 \\
\textbf{SV w/ MVHM}    &\textbf{0.832} &\textbf{0.991}  & \textbf{8.38} \\
\hline
\end{tabular}
\label{table:3Dresult}
\end{table}

\begin{table}
\centering
\caption{Results on the MHP dataset. $\uparrow$: higher is better.}
    \begin{tabular}{lc}
    \hline
&$\text{AUC}_{\text{20-50}}\uparrow$\\
\hline
Zimmermann ~\etal\cite{zimmermann2019freihand} &  0.717 \\
Cai ~\etal ~\cite{cai2018weakly} & 0.928 \\
Chen ~\etal~\cite{chen2020dggan} & 0.939\\
\textbf{Our multi-view method} &  \textbf{0.991}  \\
\hline
\end{tabular}
\label{table:sotaResult}
\vspace{-.3cm}
\end{table}

\begin{figure*}
\centering 
\begin{tabular}{cccc}
\hspace{-.6cm}
\includegraphics[width=0.25\textwidth]{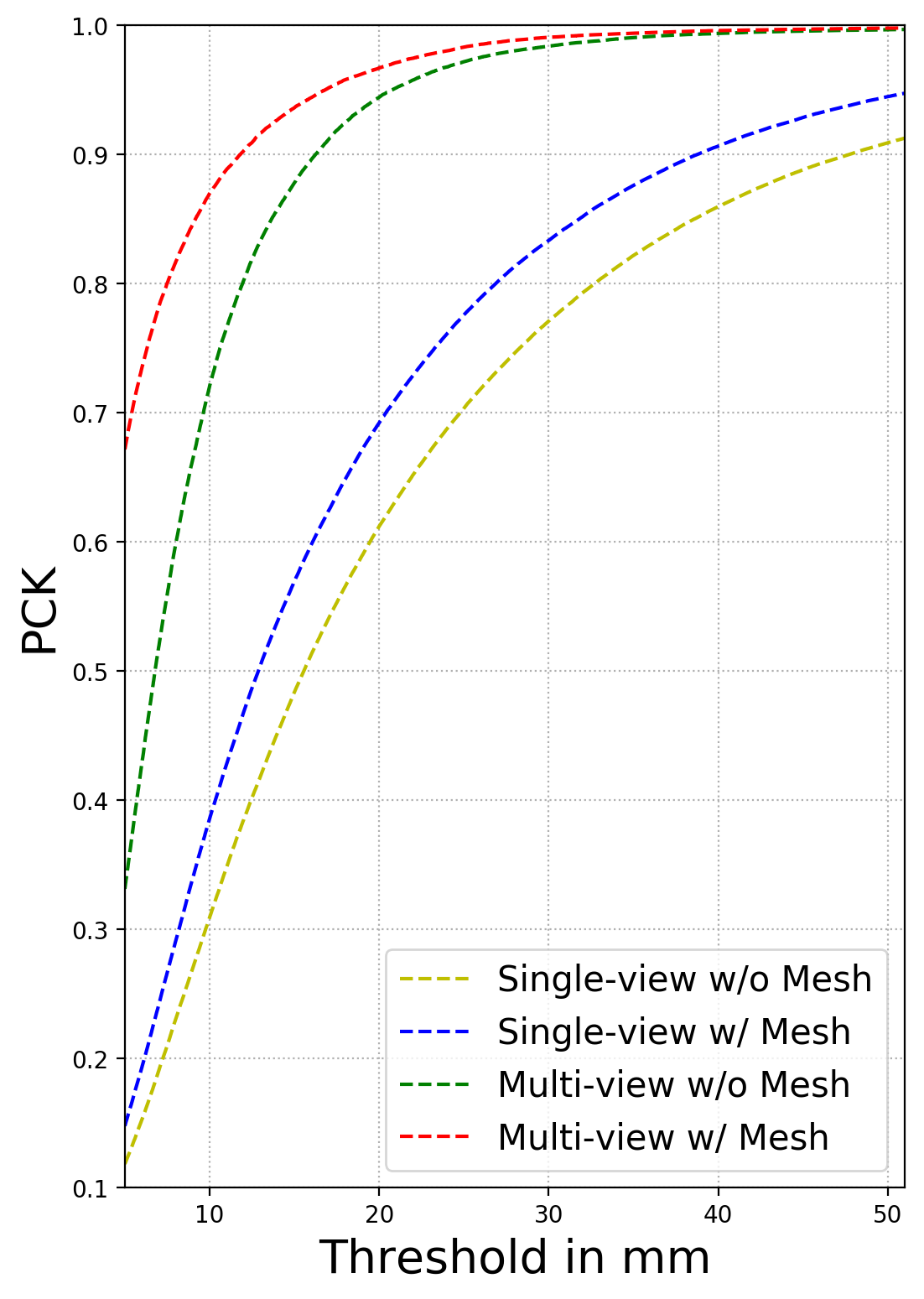}& 
\hspace{-.5cm}
\includegraphics[width=0.25\textwidth]{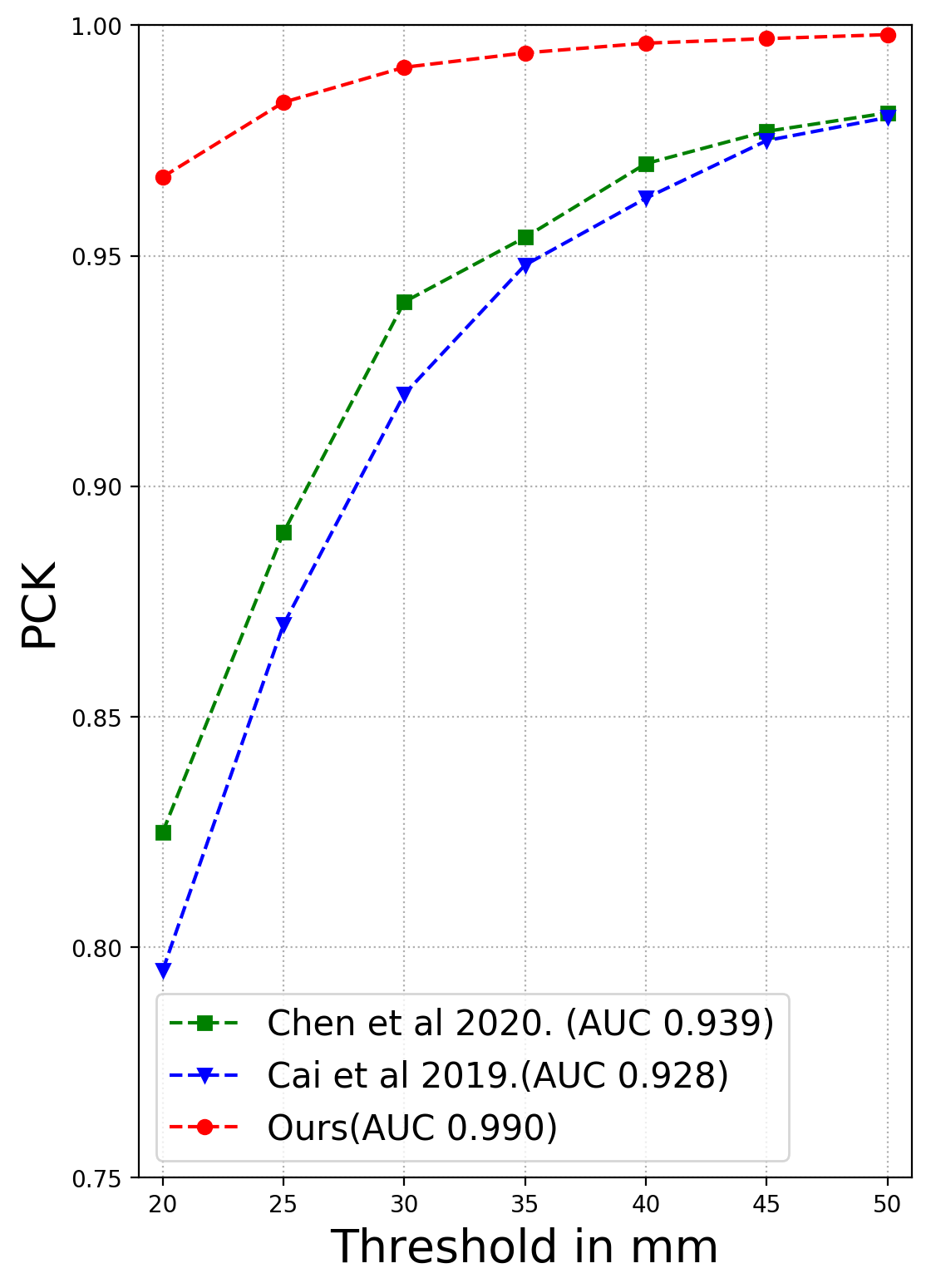}&
\hspace{-.5cm}
\includegraphics[width=0.25\textwidth]{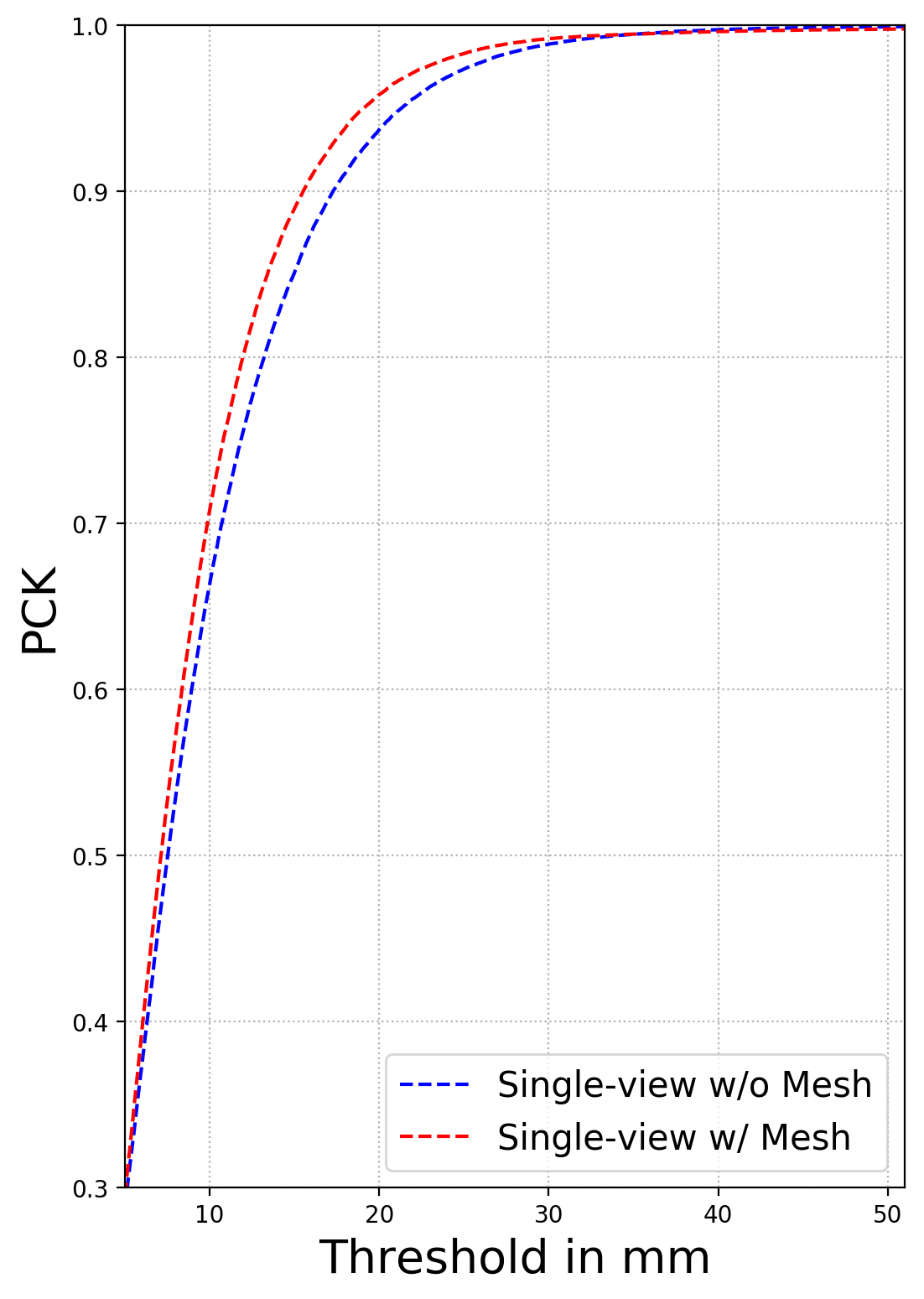}&
\hspace{-.5cm}
\includegraphics[width=0.25\textwidth]{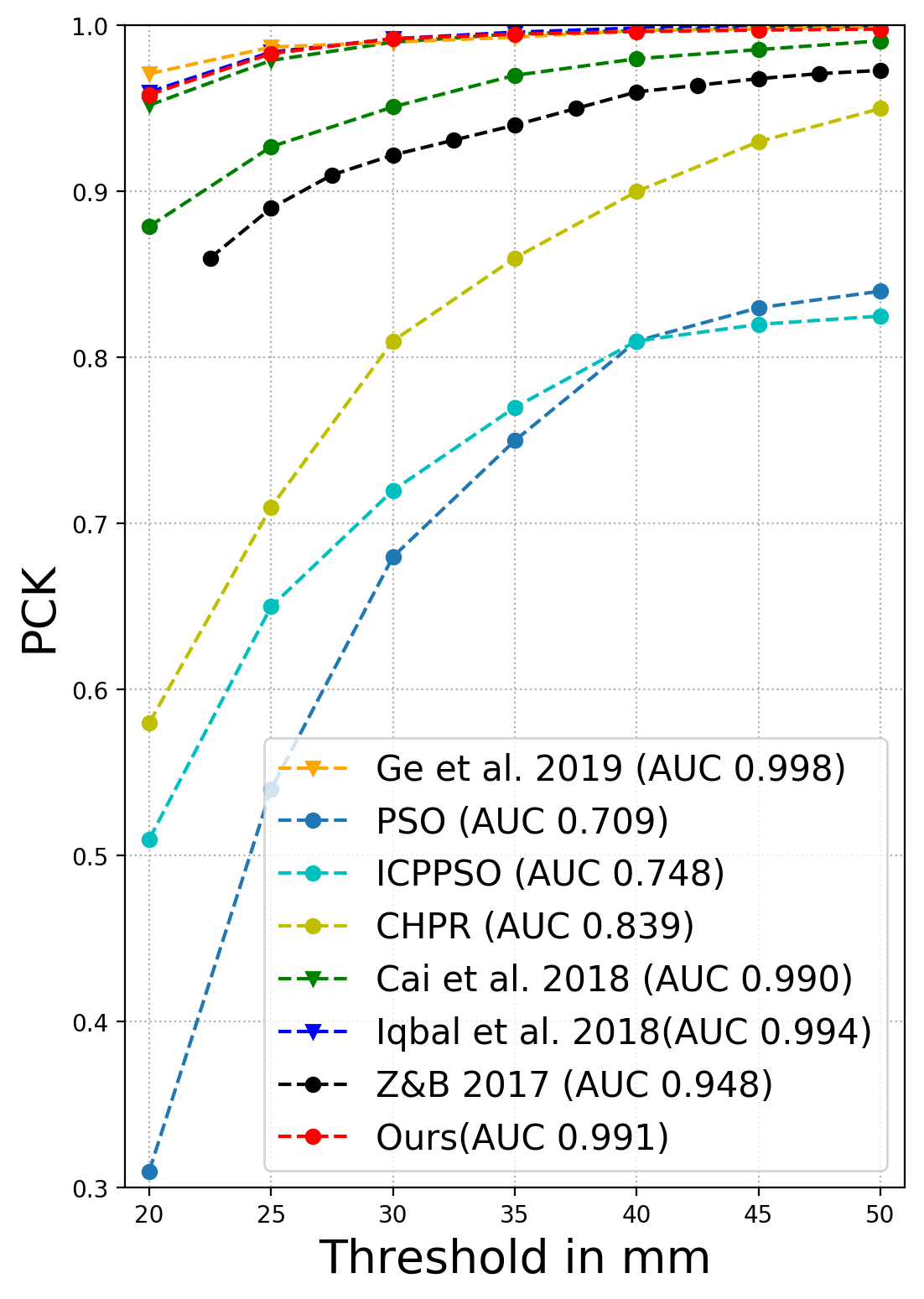}\\
(a) & (b) & (c) & (d)
\end{tabular}
\caption{
Ablation studies and comparison of the state-of-the-art methods for single-view pose estimation. (a) PCK results of different settings on the STB dataset. (b) Comparison results in PCK for the state-of-the-art methods on the STB dataset. (c) PCK results under different settings on the MHP dataset. (d) Comparison results in PCK for the state-of-the-art methods on the MHP dataset. 
}
\label{fig:result}
\vspace{-.2cm}
\end{figure*}

\subsection{Metrics}

We follow the settings from previous researches~\cite{zb2017hand,ge2019handshapepose} and adopt the {\em average end-point-error} ($\text{EPE}_{m}$), 
and the {\em area under the curve} (AUC) on the {\em percentage of correct keypoints} (PCK) within a threshold range as the metrics to evaluate model effectiveness. 
We report the performance in both AUC on PCK between 0mm and 50mm as well as between 20mm and 50mm.
\vspace{-0.1cm}
\subsection{Implementation Details}

We implement our single-view and multi-view approaches in Python with PyTorch.
In the training phase, we set the batch size as $8$, and use the Adam solver with an initial learning rate 0.01. 
Both models are trained on a server with four GeForce GTX 1080-Ti GPUs.

%
When training the single-view network, we use a multi-stage training strategy.
In the first stage, we train our 2D evidence network with the heatmap loss $\bm{\mathcal{L}}_{h}$.
In the second stage, we fix the weights of the 2D evidence network and train the mesh network with mesh loss $\bm{\mathcal{L}}_{m}$. 
In the third stage, we fix the weights of both of the 2D evidence network and mesh network, and focus on training the joint depth network with loss $\bm{\mathcal{L}}_{d}$. 
In the final stage, the whole network is optimized end-to-end.

For training the multi-view network, we apply the same multi-stage training strategy. 
In the first stage, we use the pre-trained weights from the single-view network for initializing the 3D hand single-view network, and keep this part fixed for training the 3D hand fusion network. 
In the second stage, we activate both networks and fine-tune the whole network architecture in an end-to-end manner.

\section{Experimental Results}

\subsection{Multi-view task}
To evaluate the effectiveness of the proposed multi-view method,
we compare our single-view method with our multi-view method on the MHP dataset under the setting of with or without using data from the MVHM dataset.
Table~\ref{table:3Dresult} and Figure~\ref{fig:result}(a) show that utilizing the multi-view information from the MHP dataset itself boosts the testing performance in $\text{AUC}_\text{0-50}$, $\text{AUC}_{\text{20-50}}$, and $\text{EPE}_m$ by large margins, \ie, 0.218, 0.183, and 13.80mm respectively. 
When additional data from the MVHM dataset are used, substantial performance gains are achieved, which reveals the effectiveness of using the collected MVHM dataset for training.
%

Three current state-of-the-art methods are chosen for comparing with our method on the MHP dataset, including Zimmermann~\etal~\cite{zimmermann2019freihand} (0.717 in $\text{AUC}_{\text{20-50}}$ ), Cai~\etal~\cite{cai2018weakly} (0.928 in $\text{AUC}_{\text{20-50}}$)\footnote{Cai~\etal~\cite{cai2018weakly} do not report the results in their paper. Here we report the re-implementaition results by Chen~\etal~\cite{chen2020dggan}.}, and Chen~\etal~\cite{chen2020dggan} (0.939 in $\text{AUC}_{\text{20-50}}$). 
Zimmermann~\etal~\cite{zimmermann2019freihand} just report the numerical result so we include their result in Table \ref{table:sotaResult} and does not show it in Figure \ref{fig:result}(b). 
Our multi-view method achieves the performance of 0.990 in $\text{AUC}_{\text{20-50}}$, outperforming these competing methods by a large margin.
This experiment shows that both the proposed multi-view method and the established MVHM dataset are beneficial and can work together to get the new state-of-the-art performance on the MHP dataset.

\subsection{Single-view task}

To further validate the effectiveness of the generated mesh dataset MVHM in addition to multi-view methods, we also conduct the following experiments for comparison on single-view methods.
We compare the results when models are trained solely on the MHP/STB datasets and trained on the MHP/STB datasets together with the MVHM dataset. 
Table~\ref{table:3Dresult}, Figure~\ref{fig:result}(a) and Figure~\ref{fig:result}(c) show, on both MHP and STB datasets, adding the mesh data greatly enhances the performance by granting a model the ability to capture the mesh-level features, therefore leading to better results.

We select seven powerful and recently published methods for comparison with the proposed method, including PSO~\cite{boukhayma20193d}, ICPPSO~\cite{chentaChen2018GeneratingRTgan}, CHPR~\cite{zhang20163d}, ~Iqbal~\etal~\cite{iqbal2018hand}, Cai~\etal~\cite{cai2018weakly}, Zimmermann and Brox~\cite{zb2017hand}, and Ge ~\etal~\cite{ge2019handshapepose}. 
The AUC curves are plotted in Figure~\ref{fig:result}(d). 
Ge~\etal~\cite{ge2019handshapepose} also utilize an additional dataset to train their model and got the STOA result, which demonstrates the effectiveness of their mesh dataset. 
Besides, they introduce more complicated mesh metrics like the surface norm loss.
Iqbal~\etal~\cite{iqbal2018hand} and Cai~\etal~\cite{cai2018weakly} leverage additional depth-map information to derive their models, and achieve good results.
As a multi-view approach without complicated components, our method is on par with methods by Ge~\etal~\cite{ge2019handshapepose} and Iqbal~\etal~\cite{iqbal2018hand} while outperforms most of them on single-view tasks. 

\section{Conclusions}
Estimating 3D hand poses from monocular images is an ill-posed problem due to its depth ambiguity. 
Nevertheless, multi-view images could make up the deficiency. 
To this end, we build a multi-view mesh hand dataset, MVHM, to enable training 3D pose estimators with mesh supervision.
We present a multi-view method that effectively fuses single-view predictions.
When testing on the real-world multi-view dataset MHP, our multi-view method with the aid of the MVHM dataset achieves the state-of-the-art performance.
\label{sec:conclu}
\vspace{-.3cm}
\paragraph{Acknowledgement.}
The authors thank Kun Han for helping generate the dataset images used in this work. This work was supported in part by the Ministry of Science and Technology (MOST) under grants MOST 107-2628-E-009-007-MY3, MOST 109-2634-F-007-013, and MOST 109-2221-E-009-113-MY3, and by Qualcomm through a Taiwan University Research Collaboration Project.

\clearpage

{\small
\bibliographystyle{ieee_fullname}
\bibliography{egbib}
}

\end{document}